%% file: main.tex
\documentclass[10pt,twocolumn,letterpaper]{article}

\usepackage[pagenumbers]{cvpr} 

\usepackage{graphicx}
\usepackage{amsmath}
\usepackage{amssymb}
\usepackage{booktabs}
\usepackage{nicefrac}
\usepackage{float}
\usepackage[accsupp]{axessibility}  

\usepackage[table]{xcolor}
\definecolor{blue}{HTML}{0055cc}
\definecolor{red}{HTML}{cc1100}   
\definecolor{orange}{HTML}{cc7700}
\usepackage[pagebackref,breaklinks, colorlinks,linkcolor=red,citecolor=blue]{hyperref}
\usepackage[capitalize]{cleveref}
\crefname{section}{Sec.}{Secs.}
\Crefname{section}{Section}{Sections}
\Crefname{table}{Table}{Tables}
\crefname{table}{Tab.}{Tabs.}
\def\cvprPaperID{0000} 
\def\confName{CVPR}
\def\confYear{2023}
\input{res/shortcuts.tex}

\input{res/math_commands.tex}

\begin{document}
 
\title{Hierarchical Dense Correlation Distillation for Few-Shot Segmentation}

\author{
Bohao Peng\textsuperscript{1}, Zhuotao Tian\textsuperscript{2}\thanks{Corresponding Author},
Xiaoyang Wu\textsuperscript{3}, Chenyao Wang\textsuperscript{1}, Shu Liu\textsuperscript{2},
Jingyong Su\textsuperscript{4}, Jiaya Jia\textsuperscript{1,2} \\ \\ 
\textsuperscript{1}The Chinese University of Hong Kong\quad 
\textsuperscript{2}SmartMore \\
\textsuperscript{3}The University of Hong Kong \quad
\textsuperscript{4}Harbin Institute of Technology, Shenzhen \\
}
 
\maketitle 

\begin{abstract}
Few-shot semantic segmentation (FSS) aims to form class-agnostic models segmenting unseen classes with only a handful of annotations. Previous methods limited to the semantic feature and prototype representation suffer from coarse segmentation granularity and train-set overfitting. In this work, we design Hierarchically Decoupled Matching Network (HDMNet) mining pixel-level support correlation based on the transformer architecture. The self-attention modules are used to assist in establishing hierarchical dense features, as a means to accomplish the cascade matching between query and support features. Moreover, we propose a matching module to reduce train-set overfitting and introduce correlation distillation leveraging semantic correspondence from coarse resolution to boost fine-grained segmentation. Our method performs decently in experiments. We achieve $50.0\%$ mIoU on \coco~dataset one-shot setting and $56.0\%$ on five-shot segmentation, respectively. The code is available on the project website\footnote{https://github.com/Pbihao/HDMNet}. 
\end{abstract} 

\section{Introduction}
\label{sec:intro}

\begin{figure}
    \centering
    \includegraphics[width=0.475\textwidth]{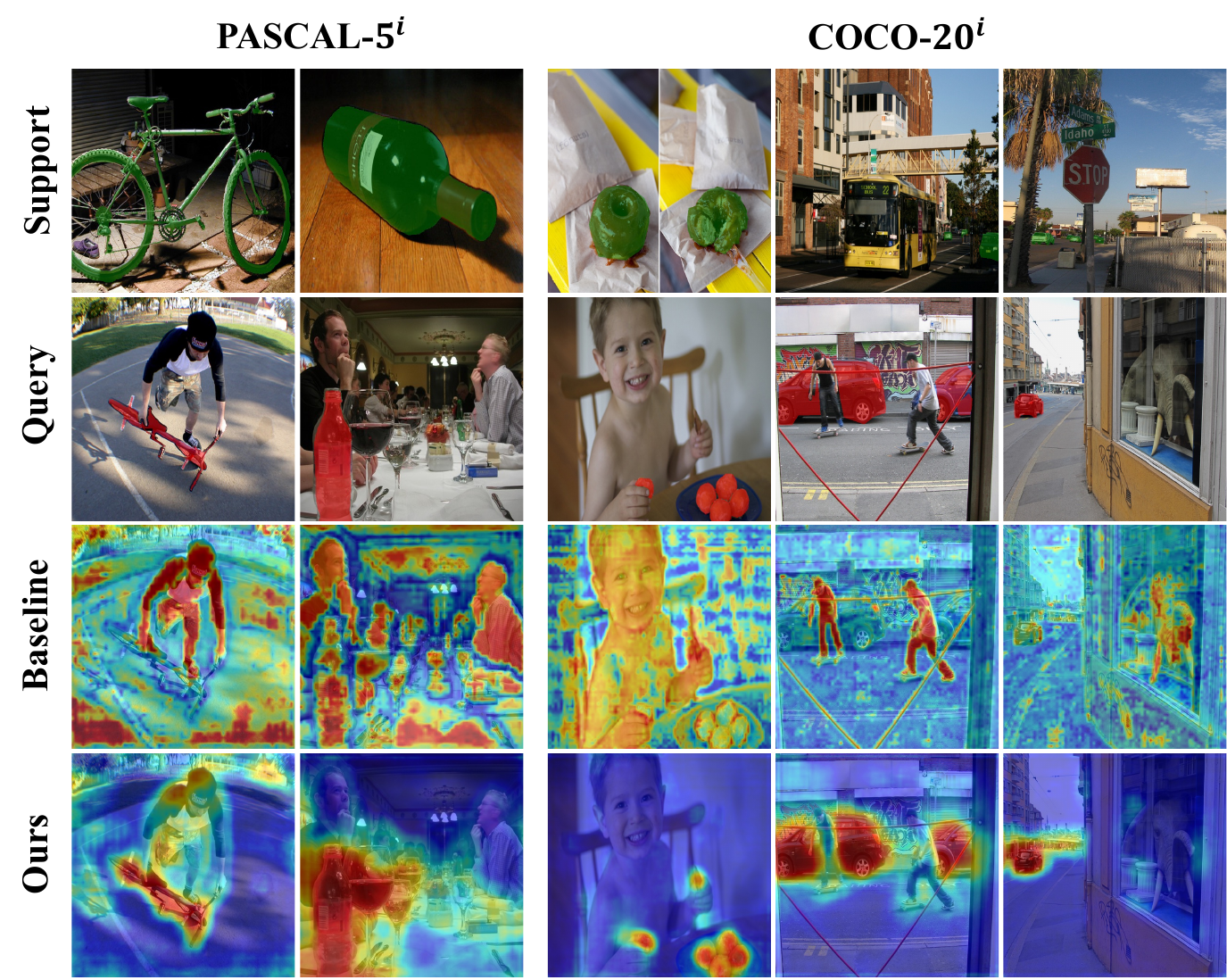}
    \caption{Activation maps of the correlation values on both $\text{PASCAL-}5^i$~\cite{shaban2017oneshot} and $\text{COCO-}20^i$~\cite{nguyen2019featureweight}. The baseline is prone to give high activation values to the categories sufficiently witnessed during training, such as the ``People'' class, even with other support annotations. Then we convert it to the hierarchically decoupled matching structure and adopt correlation map distillation to mine inner-class correlation.}
    \label{fig:correlation_visualization}
   \vspace{-0.2in}
\end{figure}
 
Semantic segmentation tasks~\cite{long2015fcn,zhao2017pspnet,tian2023cac,zhang2022mediseg,tian2019lase} have made tremendous progress in recent years, benefiting from the rapid development of deep learning~\cite{He2016resnet,reslt,simonyan2014very,lai2022decouplenet,cui2022generalized}. However, most existing deep networks are not scalable to previously unseen classes and rely on annotated datasets to achieve satisfying performance. Data collection and annotation cost much time and resources, especially for dense prediction tasks~\cite{lai2021cac,jiang2021semi,cui2022region}.

Few-shot learning~\cite{vanschoren2018meta,wang2020generalizing,snell2017prototypical} has been introduced into semantic segmentation~\cite{dong2018protypelearn,tian2020prior} to build class-agnostic models quickly adapting to novel classes. Typically, few-shot segmentation (FSS) divides the input into the query and support sets~\cite{zhang2019pyramid,dong2018protypelearn,zhang2019canet,zhang2020sg-one} following the episode paradigm~\cite{vinyals2016matching}. It segments the query targets conditioned on the semantic clues from the support annotations with meta-learning~\cite{snell2017prototypical,vanschoren2018meta} or feature matching~\cite{min2021hsnet,vinyals2016matching,zhang2021cycle}.

\begin{figure}
    \centering
    \includegraphics[width=0.43\textwidth]{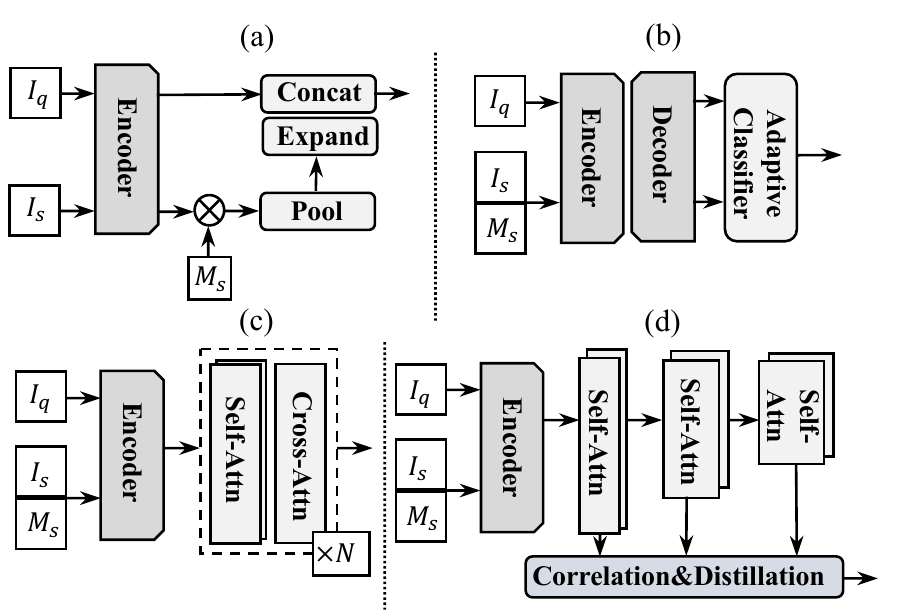}
    \caption{Illustration of different few-shot segmentation frameworks. (a) Prototype-based method. (b) Adaptive-classifier method. (c) Feature matching with transformer architecture. (d) Our Hierarchically Decoupled Matching Network (HDMNet) with correlation map distillation.}
   \vspace{-0.3in}
    \label{fig:architecture_comparison}
\end{figure}

Previous few-shot learning methods may still suffer from coarse segmentation granularity and train-set overfitting~\cite{tian2020prior} issues. As shown in~\figref{fig:correlation_visualization}, ``people'' is the base class that has been sufficiently witnessed during training. But the model is still prone to yield high activation to ``people'' instead of more related novel classes with the support samples, producing inferior results. This issue stems from framework design, as illustrated in \figref{fig:architecture_comparison}.
Concretely, prototype-based~\cite{tian2020prior,wang2019panet} and adaptive-classifier methods~\cite{lu2021cwt,boudiaf2021RePRI} aim at distinguishing different categories with global class-wise characteristics. It is challenging to compute the correspondence of different components between query and support objects for the dense prediction tasks. In contrast, matching-based methods~\cite{zhang2021cycle} mine pixel-level correlation but may heavily rely on class-specific features and cause overfitting and weak generalization.

To address these issues, we propose Hierarchically Decoupled Matching Network (HDMNet) with correlation map distillation for better mining pixel-level support correspondences. HDMNet extends transformer architecture~\cite{vaswani2017attention,dosovitskiy2020vit,xie2021segformer} to construct the feature pyramid and performs dense matching. Previous transformer-based methods~\cite{zhang2021cycle,sun2021loftr} adopt the self-attention layer to parse features and then feed query and support features to the cross-attention layer for pattern matching, as illustrated in~\figref{fig:architecture_comparison}(c). This process stacks the self- and cross-attention layers multiple times, mixes separated embedding features, and accidentally causes unnecessary information interference. 

In this paper, we decouple the feature parsing and matching process in a hierarchical paradigm and design a new matching module based on correlation and distillation. This correlation mechanism calculates pixel-level correspondence without directly relying on the semantic-specific features, alleviating the train-set overfitting problem. Further, we introduce correlation map distillation~\cite{hinton2015distilling,zhang2019selfdistill} that encourages the shallow layers to approximate the semantic correlation of deeper layers to make the former more aware of the context for high-quality prediction.
 
Our contribution is the following. 1) We extend the transformer to hierarchical parsing and feature matching for few-shot semantic segmentation, with a new matching module reducing overfitting. 2) We propose correlation map distillation leveraging soft correspondence under multi-level and multi-scale structure. 3) We achieve new state-of-the-art results on standard benchmark of $\text{COCO-}20^i$ and $\text{PASCAL-}5^i$ without compromising efficiency.   
 
\section{Related Work}
\label{sec:related}

\begin{figure*}[!ht]
    \centering
    \includegraphics[width=\textwidth]{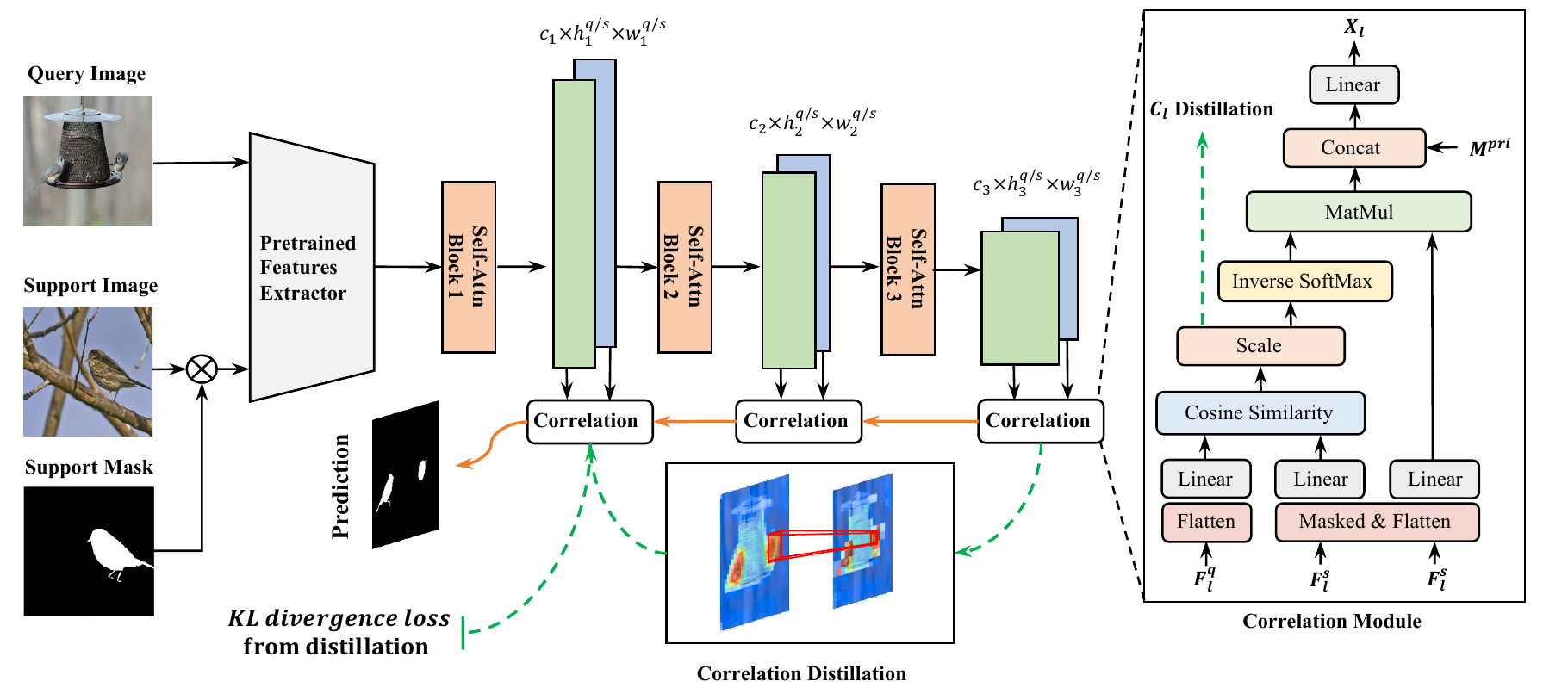}
    \caption{Schematic overview of the proposed few-shot semantic segmentation model. $h_l^{\nicefrac{q}{s}}=\frac{H^{\nicefrac{q}{s}}}{2^{l+2}}$, $w_l^{\nicefrac{q}{s}}=\frac{W^{\nicefrac{q}{s}}}{2^{l+2}}$ indicate the height and width of the $l$-th stage features, and $H^{\nicefrac{q}{s}}$ and $W^{\nicefrac{q}{s}}$ are the height and width of the input query and support images. $c_l$ represents the channels and $c_{l+1}>c_{l}$.}
    \label{fig:architecture}
   \vspace{-0.2in}
\end{figure*}

\paragraph{Few-shot semantic segmentation.} Few-Shot Semantic Segmentation (FSS)~\cite{fan2020fgn,dong2018protypelearn,gairola2020simpropnet,tian2020prior} predicts dense masks for novel classes with only a few annotations. Previous approaches following metric learning~\cite{dong2018protypelearn,wang2019panet,sung2018relationnetwork,tian2020prior} can be divided into prototype- and matching-based methods. Motivated by PrototypicalNet~\cite{snell2017prototypical} for few-shot learning, the prevalent FSS models~\cite{wang2019panet,dong2018protypelearn,fan2022self,tian2022generalized,luo2021pfenet++} utilize prototypes for specific-class representation. Recent work~\cite{zhang2019regionbased,li2021adaptive,zhang2021cycle} points out that a single prototype has a limitation to cover all regions of an object, especially for pixel-wise dense segmentation tasks. To remedy this problem, methods of~\cite{zhang2019regionbased,li2021adaptive} use EM and cluster algorithms to generate multiple prototypes for different parts of the objects. Compared with prototype-based methods, matching-based ones~\cite{vinyals2016matching,min2021hsnet,zhang2021cycle, lu2021cwt} are based on intuition to mine dense correspondence between the query images and support annotations. They utilize pixel-level features and supplement more detailed support context. 

\vspace{-0.2in}
\paragraph{Transformer.} Transformer was first introduced in Natural Language Processing (NLP)~\cite{vaswani2017attention,devlin2018bert}. In computer vision, ViT~\cite{dosovitskiy2020vit} treats an image as a patch sequence and demonstrates that pure transformer architecture can achieve state-of-the-art image classification. Recent work explores combining few-shot semantic segmentation and transformer architecture~\cite{sun2021loftr,li2021disparityestimation}. In~\cite{lu2021cwt}, the classifier weight transformer adapts the classifier's weights to address the intra-class variation issue. CyCTR~\cite{zhang2021cycle} is a cycle-consistent transformer by generating query and key sequences from the query and support set, respectively. Transformer architecture helps FSS transcend the limitation of semantic-level prototypes and leverage pixel-wise alignment. Previous transformer-based methods are still difficult  to handle noise interference and over-fitting. We, in this paper, propose a new transformer structure decoupling the downsampling and matching processes and design the matching module constructed on correlation mechanism and distillation.

\vspace{-0.2in}
\paragraph{Knowledge distillation.} Knowledge distillation (KD)~\cite{hinton2015distilling} was widely used in model compression. Large models typically have higher knowledge capacity. In contrast, small models have fewer parameters, better efficiency, and lower cost. KD attempts to transfer learned knowledge from the large model (a.k.a. the teacher) to another light model (a.k.a. the student) with tolerable loss in performance. Method of~\cite{zhang2019selfdistill} processes a self-distillation framework to distill knowledge within itself to improve model accuracy. Self-distillation divides the model into multiple sections and transfers knowledge from deeper portions to shallow ones. Knowledge distillation is also used for extracting semantic information and mining inner correlation. STEGO~\cite{hamilton2022stego} shows that even unsupervised deep network features have correlation patterns consistent with true labels. STEGO framework applies feature correlation map distillation to excavate the intrinsic semantic correlation at pixel level. APD~\cite{apd} leveraged distillation to tackle semantic segmentation. Motivated by unsupervised semantic segmentation~\cite{hamilton2022stego}, we design the correlation mechanism for class-agnostic feature matching and propose multi-level multi-scale correlation map distillation to transfer relation between the query and support set from deep sections to the shallow ones.

\section{Task Definition and Our Method}
\label{sec:task}

Few-shot segmentation is to train segmentation for novel objects with only a few annotated support images. In definition, the model is trained on ${\mD_{train}}$ and is evaluated on $\mD_{test}$. Suppose the category sets in $\mD_{train}$ and  $\mD_{test}$ are $\mC_{train}$ and $\mC_{test}$ respectively. There is no intersection between the training and testing sets, \ie, $\mC_{train} \cap \mC_{test} = \emptyset$. Following previous work \cite{shaban2017one, tian2020prior, wang2019panet, zhang2019canet}, episodes are applied to both train set $\mD_{train}$ and test set $\mD_{test}$. 
  
Each episode is composed of a query set $\mQ = \{(\mI^q, \mM^q)\}$ and a support set $\mS = \{(\mI^s_i, \mM^s_i)\}_{i=1}^K$ with the same class $c$ , where $\mI^q,\mI^s \in \mathbb{R}^{H\times W\times 3} $ represent the RGB images and $\mM^q,\mM^s \in \mathbb{R}^{H\times W}$ denote their binary masks. Both the query masks $\mM^q$ and the support masks $\mM^s$ are used during the training process, while only the support masks $\mM^s$ are accessible in testing. Since the model parameters are fixed and require no optimization for novel categories during testing, the model is trained to leverage the semantic clues provided by the support set to locate the regions of interest on the query images. 

\subsection{Hierarchically Decoupled Matching Network}
\label{sec:method}

Given the query set $\mQ=\{(\mI^q, \mM^q)\}$ and the support set $\mS=\{(\mI^s_i, \mM^s_i)\}_{i=1}^K$, HDMNet adopts a parameter-fixed encoder to extract rich features of the query and support images, following~\cite{zhang2019canet,tian2020prior}. The difference is on the design of new decoder to yield predictions on the query images by decently leveraging pixel-level feature matching between the query and support sets. 

Overview of the pipeline is shown in~\figref{fig:architecture}. In the following, we start by putting forward the basic structure of HDMNet in~\secref{sec:MSSMStructure}, followed by the introduction regarding the proposed correlation calculation strategy in~\secref{sec:corr-mechanism}. Then, the inter-stage correlation map distillation is presented in~\secref{sec:correlation_distillation}. Finally, in~\secref{sec:kshotsetting}, we instantiate the way to extend to the K-shot setting.

\subsection{Overview of the Architecture}
\label{sec:MSSMStructure}

\begin{figure}
    \centering
    \includegraphics[width=0.4\textwidth]{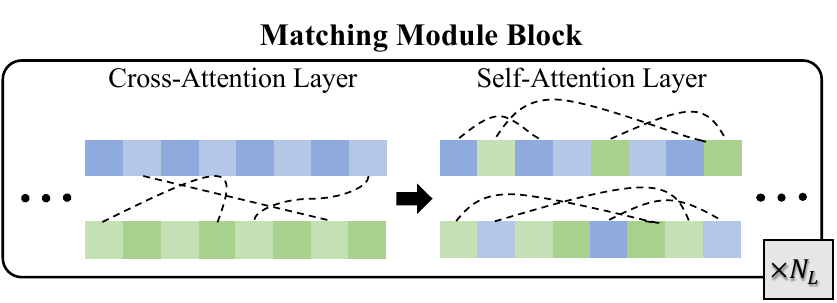}
    \caption{Feature Matching with directly stacking cross-attention layers and self-attention layers. It's intuitive to notice that the cross-attention layer mixes the query and support features, destroying the purity of parsing and matching consistency.}
    \vspace{-.2in}
    \label{fig:directly_stack}
\end{figure}

\paragraph{Motivation.} In previous matching-based methods with transformer architecture, the self-attention and cross-attention layers are interleaved for multiple times for feature parsing and pattern matching respectively~\cite{zhang2021cycle,sun2021loftr} as shown in~\figref{fig:directly_stack}. We note that the cross-attention layers accomplish mutual message exchange between the query and support features. Objects in the background of the query sample may also correlate with the target in the support sample. Thus, they can be enriched with support information. With this finding, the necessary support information may be accumulated to the distracters via multiple stacked cross-attention and self-attention layers, making the decoder harder to distinguish among them. 

To ensure the purity of the sequential features and consistency of pattern matching, we propose a new hierarchically matching structure decoupling the down-sampling and matching processes, where only independent self-layers are adopted to build hierarchical features.  

\vspace{-0.1in}
\paragraph{Decoupled downsampling and matching.}
First, the query and support features extracted from the backbone are independently sent to sequential transformer blocks with only the self-attention layers to fully exploit self-correlation within the support and query features. We note that the down-sampling layers are inserted between blocks to establish a hierarchical structure that may assist in mining the inter-scale correlations.

Then, the intermediate feature maps of $L$ stages are collected, i.e., $\{\mF^q_l\}_{l=1}^L$ and $\{\mF^s_l\}_{l=1}^L$. Assume $\{\mF^q_l\}$ and $\{\mF^s_l\}$ have the same spatial size $[c_l\times h_l^{\nicefrac{q}{s}}\times w_l^{\nicefrac{q}{s}}]$ for simplicity's sake. $$h_l^{\nicefrac{q}{s}}=\frac{H^{\nicefrac{q}{s}}}{2^{l+2}},\quad w_l^{\nicefrac{q}{s}}=\frac{W^{\nicefrac{q}{s}}}{2^{l+2}},$$ $l$ is the stage index, and $c_l$ denotes the feature channel number. 
Finally, $\{\mF^q_l\}_{l=1}^L$ and $\{\mF^s_l\}_{l=1}^L$ are used to yield correlations $\{\mC_l\in \R^{h_l^qw_l^q\times h_l^sw_l^s}\}_{l=1}^L$ and enriched query features $\{\mX_l\in\R^{c_l\times h_l^q\times w_l^q}\}_{l=1}^L$. Detailed formulations are elaborated later in Eqs. \eqref{eq:corr_map} and \eqref{eq:corr_mechanism} in Sec.~\ref{sec:corr-mechanism}.

\vspace{-0.1in}
\paragraph{Coarse-grained to fine-grained decoder.} HDMNet incorporates a simple decoder to predict the final mask for the query image with the hierarchically enriched features $\{\mX_l\in\R^{c_l\times h_l^q\times w_l^q}\}_{l=1}^L$ in a coarse-to-fine manner. Specifically, the coarse-grained features $\mX'_{l+1}$ are scaled up to have the same spatial size as the fine-grained one $\mX'_{l}$. Then an MLP layer is adopted to fuse them with a residual connection, written as
\begin{equation}
    \mX'_l=\text{ReLU}(\text{MLP}(\mX_l+\zeta_l(\mX'_{l+1})))+\zeta_l(\mX'_{l+1}),  
    \label{eq:decoder} 
\end{equation}
where $l$ indicates the hierarchical stage, and $\zeta_l:\mathbb{R}^{H\times W}\mapsto\mathbb{R}^{h_l\times w_l}$ denotes the bilinear-interpolation resize function fitting the input size to that of the output. Finally, we apply a convolution layer with $1\times 1$ kernel size to $\mX'_1$ followed by a bilinear up-sampling layer to predict the query mask $\mM^{out}\in\mathbb{R}^{H\times W}$.

\begin{figure}
    \centering
    \includegraphics[width=0.46\textwidth]{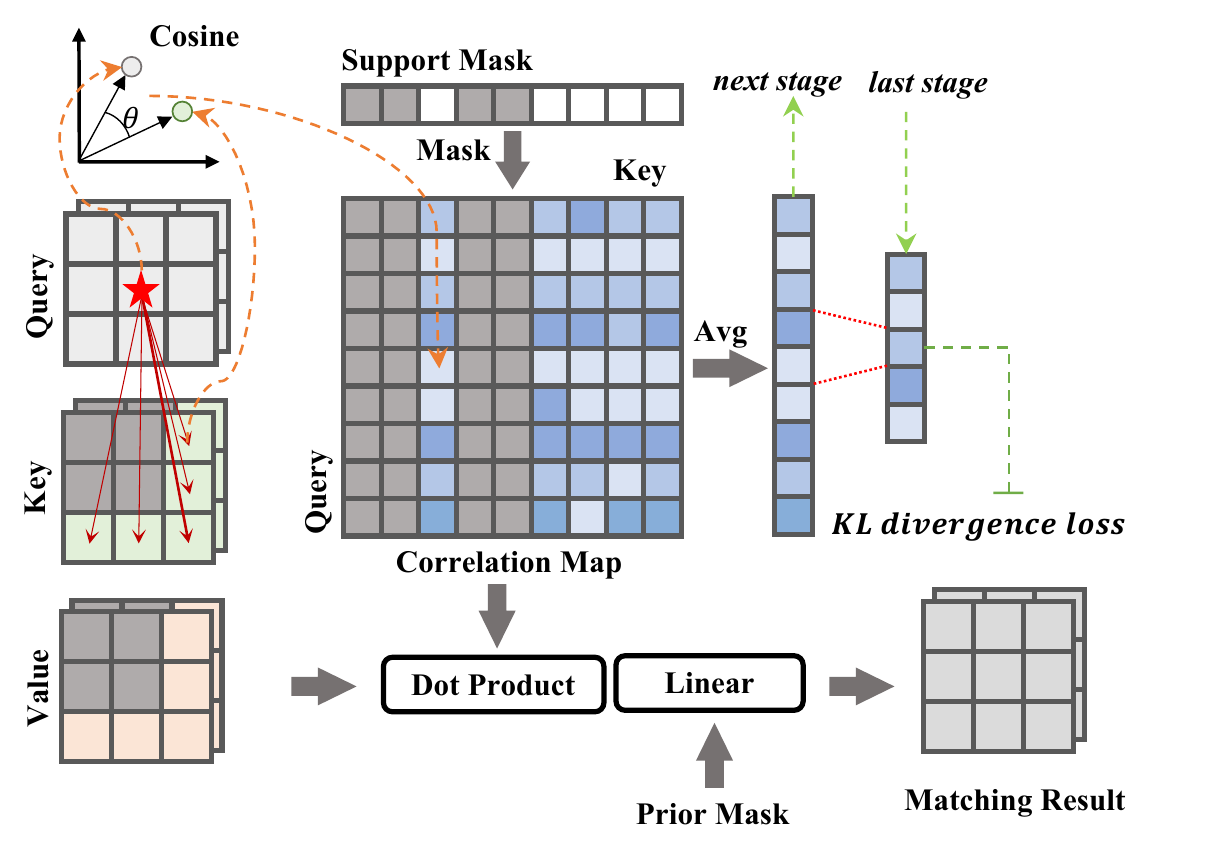}
    \caption{Illustration of our proposed matching module based on correlation mechanism and distillation.}
    \vspace{-.2in}
    \label{fig:matching_module}
\end{figure}

\subsection{Matching Module} 
\label{sec:corr-mechanism}

\paragraph{Motivation.}
Previous matching-based methods~\cite{lu2021cwt,zhang2021cycle}  directly adopt the cross-attention as the matching module by generating the query and key features from the query and support sets respectively. However, we observe that it leads to overfitting and weak generalization. This issue could be attributed to the fact that the models are more likely to rely on class-specific features to optimize the training objectives. To alleviate this issue, we propose a matching module illustrated in \figref{fig:matching_module}.

\vspace{-0.1in}
\paragraph{Attention.}
\label{sec:attention_function}
Following the general form~\cite{vaswani2017attention,xie2021segformer}, the critical element of the transformer block is  the dot-product attention layer, formulated as 
\begin{equation}
    \text{Attn}(\mQ,\mK,\mV) = \text{softmax}(\frac{\mQ\mK^T}{\sqrt{d}})\mV,
    \label{eq:attention_formulation}
\end{equation}
where $[\mQ;\mK;\mV]=[\mW^q\mF^q;\mW^k\mF^{s};\mW^v\mF^{s}]$, in which $\mF^q$ and $\mF^{s}$ denote the query and support features respectively, $\mW^q,\mW^k,\mW^v\in\mathbb{R}^{d\times d}$ are the learnable parameters, $d$ is the hidden dimension. 

The cross-attention layer takes essential support information from $\mV$, conditioned on the query-support correlation between $\mQ$ and $\mK$. When $\mF_q=\mF_{s}$, it functions as a self-attention layer for relating different positions within either the support or query input features.

\vspace{-0.1in}
\paragraph{Our correlation mechanism.}
\label{sec:correlation_mechanism}  
Our designed matching module based on the correlation mechanism retrieves the most relevant regions with high cosine similarity and fuses the high-level prior mask generated as that in~\cite{tian2020prior}. Given the query features $F^q$ and the support features $F^s$, we first transform the input features by
\begin{equation}
\begin{aligned}
    \hat{\mF}^q &= \varphi(\mF^q),\\
    \hat{\mF}^s &= \varphi(\mF^s\odot\mM^s),
\end{aligned}
\end{equation}
where $\odot$ is Hadamard product, $\varphi:\mathbb{R}^{c\times h\times w}\mapsto\mathbb{R}^{hw\times c}$ refers to the reshape function, and $\mM^s$ denotes the support mask. To mitigate the risk of overfitting the category-specific information brought by the feature norms, we measure the cosine similarities of the inner product angle, instead of performing dot product, to calculate the correlation map as $C\in\mathbb{R}^{h^qw^q\times h^sw^s}$ as
\begin{equation}
    \mC=\frac{\langle \mW^q\hat{\mF}^q,\mW^k\hat{\mF}^s\rangle }{\left\lVert \mW^q\hat{\mF}^q\right\rVert \left\lVert \mW^k\hat{\mF}^s\right\rVert t},
    \label{eq:corr_map}
\end{equation}
where $\mW^q,\mW^k\in\mathbb{R}^{c\times c}$ denote the learnable parameters, $\left\lVert\cdot\right\rVert$ indicates $L^2$ norm, and $t$ is a hyperparameter to control the distribution range, empirically set to 0.1 in all experiments. Inspired by~\cite{rocco2018neighbourhood,smith2017offline}, we propose the inverse softmax layer that normalizes the correlation matrix along the query axis since we only retrieve the interested region of the query set as
\begin{equation}
    \hat{\mC}(i,j)=\frac{\text{exp}(\mC(i,j))}{\sum_{k=1}^{h_l^qw_l^q}\text{exp}(\mC(k, j))}.
\end{equation}

Finally, we introduce the prior mask $\mM^{pri}\in\mathbb{R}^{h^q\times w^q}$ calculated the same as~\cite{tian2020prior} by concatenating it with correspondence scores along the channel dimension to generate the matching results of
\begin{equation}
    \mX=\mW^o([\psi(\hat{\mC}(\mW^v\hat{\mF^s})),\mM^{pri}]),
    \label{eq:corr_mechanism} 
\end{equation}
where $\mW^v\in\mathbb{R}^{c\times c}$, $\mW^o\in\mathbb{R}^{c\times (c+1)}$ denote the learnable parameters, $\hat{\mF}^s\in\mathbb{R}^{h^sw^s\times c},\mX\in\mathbb{R}^{c\times h^q\times w^q}$ are flattened support features and matching output, and $\psi:\mathbb{R}^{h^qw^q\times c}\mapsto \mathbb{R}^{c\times h^q\times w^q}$ is the reshape function. 

\subsection{Correlation Map Distillation} 
\paragraph{Motivation.} The query and support features are constructed in a hierarchical structure, used to generate multi-level multi-scale dense correlation map. High-level correspondence typically has more informative semantic cues. It only gives rough location limited by the coarse resolution, while the low stages contain more locally detailed features. We propose correlation map distillation to encourage the correlation maps of earlier stages to retain the fine-grained segmentation quality without deprecating the contextual hints.

In addition, compared to the one-hot labels in ground-truth annotation, the soft targets adopted during distillation can reveal the distributions among all categories, providing extra semantic cues regarding the inter-class relation. It facilitates the shallower ones to capture broader context information. On the contrary, the ground-truth annotation may not provide such information and may even let the model over-fit noises contained in the one-hot labels.

\vspace{-0.1in}
\paragraph{Distillation formulation.} 
\label{sec:correlation_distillation}
Eq.\eqref{eq:corr_map} calculates the correlation maps $\{\mC_l\in \R^{h_l^qw_l^q\times h_l^sw_l^s}\}_{l=1}^L$ for the query and support features. We reorganize $\mC_l$ with mean average and filter the irrelevant information by the support mask $M^s$ as
\begin{equation}
    \mC'_{l}(i) = \frac{\sum_{j=1}^{h_l^sw_l^s}\mC_{l}(i,j)\cdot[\varphi\circ\zeta_l(\mM^s)(j)>0]}{\sum_{j=1}^{h_l^sw_l^s}[\varphi\circ\zeta_l(\mM^s)(j)>0]},
    \label{eq:map_reorgain}
\end{equation}
where $l$ indicates the stage, $\mC'_l\in\R^{h_l^qw^q_l}$, and $\zeta_l$ is the resize function. Given flattened correlation maps, we apply a softmax layer to perform spatial normalization among all positions as
\begin{equation}
    \hat{\mC}'_l(i)=\frac{\text{exp}({\mC}'_l(i)/T)}{\sum_{j=1}^{h_l^qw_l^q}\text{exp}({\mC}'_l(j)/T)},
    \label{eq:softmax_normalization}
\end{equation}
where $l$ indicates the stage and $T$ denotes the temperature of distillation~\cite{hinton2015distilling} set to 1. Moreover, the results regarding the temperature $T$ are shown in the supplementary file.

Then the KL (Kullback-Leibler) divergence loss is used as supervision from the teacher to student with their softmax output. The correlation maps of adjacent stages act as the teacher and student respectively, formulated as
\begin{equation}
\begin{split}    
    \mathcal{L}_{KL} &= \sum_{x\in\mathcal{X}}\phi_{t}(x)\text{log}(\frac{\phi_{t}(x)}{\phi_{s}(x)}) \\
    &= \sum\nolimits_{i=1}^{h_l^qw_l^q}\zeta_l(\hat{\mC}_{l+1})(i)\cdot\text{log}(\frac{\zeta_l(\hat{\mC}_{l+1})(i)}{\hat{\mC}_{l}(i)}),
    \label{eq:kldiv_loss}
\end{split}
\end{equation}
where $l$ indicates the stage, $\phi_t$ is the teacher model while $\phi_s$ is the student model, and $\zeta_l:\mathbb{R}^{h_{l+1}^qw_{l+1}^q}\mapsto \mathbb{R}^{h_l^qw_l^q}$ represents resizing. In particular, for the last correlation map without successor, we directly utilize the ground-truth as its teacher.

\begin{table*}[ht]
    \small
    \centering
    \begin{tabular}{l l|ccccc|ccccc}
    \specialrule{\wideline}{0pt}{0pt}
    \hline
         \multirow{2}{*}{Backbone}&\multicolumn{1}{l|}{\multirow{2}{*}{Methods}}&\multicolumn{5}{c|}{1-shot}&\multicolumn{5}{c}{5-shot}\\
         &&Fold-0&Fold-1&Fold-2&Fold-3&Mean&Fold-0&Fold-1&Fold-2&Fold-3&Mean \\
         \hline
         \multirow{6}{*}{VGG-16}
         &PANet~\cite{wang2019panet}&-&-&-&-&20.9&-&-&-&-&29.7\\
         &FWB~\cite{nguyen2019featureweight}&18.4&16.7&19.6&25.4&20.0&20.9&19.2&21.9&28.4&22.6\\
         &PRNet~\cite{liu2020prnet}&27.5&33.0&26.7&29.0&29.1&31.2&36.5&31.5&32.0&32.8\\
         &PFENet~\cite{tian2020prior}&35.4&38.1&36.8&34.7&36.3&38.2&42.5&41.8&38.9&40.4\\
         &BAM$^{\dag}$~\cite{lang2022BAM}&36.4&47.1&43.3&41.7&42.1&42.9&51.4&48.3&46.6&47.3\\
         \cline{2-12}
         &\textbf{HDMNet (Ours)}&\textbf{40.7}&\textbf{50.6}&\textbf{48.2}&\textbf{44.0}&\textbf{45.9}&\textbf{47.0}&\textbf{56.5}&\textbf{54.1}&\textbf{51.9}&\textbf{52.4}\\
    \hline
         \multirow{9}{*}{ResNet-50}
         &DCP~\cite{lang2022beyond}&40.9&43.8&42.6&38.3&41.4&45.8&49.7&43.7&46.6&46.5\\ 
         &PFENet~\cite{tian2020prior}&36.5&38.6&34.5&33.8&35.8&36.5&43.3&37.8&38.4&39.0\\
         &RPMMs~\cite{yang2020rpmm}&29.5&36.8&29.0&27.0&30.6&33.8&42.0&33.0&33.3&35.5\\
         &RePRI~\cite{boudiaf2021RePRI}&32.0&38.7&32.7&33.1&34.1&39.3&45.4&39.7&41.8&41.6\\
         &HSNet~\cite{min2021hsnet}&36.3&43.1&38.7&38.7&39.2&43.3&51.3&48.2&45.0&46.9\\
         &CAPL~\cite{tian2022generalized}&37.3&43.3&40.2&38.4&39.8&43.1&53.8&48.7&47.4&48.3\\
         &CWT~\cite{lu2021cwt}&30.3&36.6&30.5&32.2&32.4&38.5&46.7&39.4&43.2&42.0\\
         &DGPNet~\cite{johnander2022dense}&43.6&47.8&44.5&44.2&45.0&\textbf{54.7}&59.1&\textbf{56.8}&54.4&\textbf{56.2}\\
         &CyCTR~\cite{zhang2021cycle}&38.9&43.0&39.6&39.8&40.3&41.1&48.9&45.2&47.0&45.6\\
         &BAM$^{\dag}$~\cite{lang2022BAM}&39.4&49.9&46.2&45.2&45.2&43.2&53.4&49.4&48.1&48.5\\
         \cline{2-12}
         &\textbf{HDMNet (Ours)}&\textbf{43.8}&\textbf{55.3}&\textbf{51.6}&\textbf{49.4}&\textbf{50.0}&50.6&\textbf{61.6}&55.7&\textbf{56.0}&56.0\\
    \specialrule{\wideline}{0pt}{0pt}
    \hline
    \end{tabular}
    \caption{Few-shot semantic segmentation performance comparison on COCO-$20^{i}$~\cite{nguyen2019featureweight} using mIoU (\%) evaluation metric. Numbers in bold indicate the best performance. $\dag$: Reproduced following the official configuration with $10,000$ test episodes.}
    \label{tab:coco_results}
    \vspace{-4mm}
\end{table*}

\subsection{Extension to $K$-shot Setting}
\label{sec:kshotsetting}

In extension to $K$-shot $(K>1)$ setting, $K$ support images with their annotated masks $S=\{(I^s_k,M^s_k)\}_{k=1}^K$ and the query set $\{(I^q,M^q)\}$ are given. HDMNet can be quickly and easily extended to the new setting based on the matching mechanism. 

As indicated in Secs. \ref{sec:corr-mechanism} and \ref{sec:correlation_distillation}, the feature matching and distillation processes are independent of the specific size of the support features, benefiting from the correlation mechanism and preprocessing of the correlation map. To prevent information loss and ensure consistency under different settings, we concatenate the support features $\mF^s_l=\text{CONCAT}([{\mF}_{l,1}^s,{\mF}_{l,2}^s, \cdots, {\mF}_{l,K}^s])$ along the channel dimension as well as their corresponding masks directly. The decoder processing remains the same as that in the one-shot setting. 

\section{Experiments}
\label{sec:exp}

\paragraph{Datasets.} Following the setting of~\cite{tian2020prior}, we use two benchmark few-shot segmentation datasets, i.e., $\text{PASCAL-}5^i$~\cite{shaban2017oneshot} and $\text{COCO-}20^i$~\cite{nguyen2019featureweight}, to evaluate HDMNet. $\text{PASCAL-}5^i$ is built from PASCAL VOC 2012~\cite{everingham2010pascal} with additional annotations from SDS~\cite{hariharan2011sds}. It consists of $20$ classes. $\text{COCO-}20^i$ is generated from MSCOCO~\cite{lin2014mscoco} and contains $80$ categories. 

For each selected dataset, cross-validation is conducted by dividing all classes into $4$ folds evenly. We use the same split class list as that of~\cite{shaban2017oneshot,nguyen2019featureweight} on $\text{PASCAL-}5^i$ and $\text{COCO-}20^i$ respectively. Specifically, three folds serve as training data, while the remaining one is used for testing. For ensuring the performance stability and fairness for comparison, we follow~\cite{tian2020prior} randomly sample $1,000$ and $10,000$ query/support pairs for $\text{PASCAL-}5^i$ and $\text{COCO-}20^i$ testing. More analysis and discussion about datasets and test episodes are shown in supplementary materials.

\vspace{-0.1in}
\paragraph{Metrics.} We adopt the mean intersection over union (mIoU) as the main evaluation metric and foreground-background IoU (FB-IoU) as the supplement. We denote $\text{mIOU}=\nicefrac{1}{C}\sum_{i=1}^C\text{IoU}_i$, where $C$ is the number of classes in each fold, and $\text{IoU}_i$ indicates intersection-over-union for class $i$. $\text{FB-IoU}=\nicefrac{1}{2}(\text{IoU}_F+\text{IoU}_B)$, where $\text{IoU}_F$ and $\text{IoU}_B$ represent the foreground and background IoU values, ignoring the class difference and computing the average.

\begin{table*}[ht]
    \small
    \centering
    \begin{tabular}{l l|ccccc|ccccc}
    \specialrule{\wideline}{0pt}{0pt}
    \hline
         \multirow{2}{*}{Backbone}&\multicolumn{1}{l|}{\multirow{2}{*}{Methods}}&\multicolumn{5}{c|}{1-shot}&\multicolumn{5}{c}{5-shot}\\
         &&Fold-0&Fold-1&Fold-2&Fold-3&Mean&Fold-0&Fold-1&Fold-2&Fold-3&Mean \\
         \hline
         \multirow{5}{*}{VGG-16}
         &PANet~\cite{wang2019panet}&42.3&58.0&51.1&41.2&48.1&51.8&64.6&59.8&46.5&55.7\\
         &PFENet~\cite{tian2020prior}&56.9&68.2&54.4&52.4&58.0&59.0&69.1&54.8&52.9&59.0\\
         &HSNet~\cite{min2021hsnet}&59.6&65.7&59.6&54.0&59.7&64.9&69.0&64.1&58.6&64.1\\
         \cline{2-12}
         &\textbf{HDMNet (Ours)}&\textbf{64.8}&\textbf{71.4}&\textbf{67.7}&\textbf{56.4}&\textbf{65.1}&\textbf{68.1}&\textbf{73.1}&\textbf{71.8}&\textbf{64.0}&\textbf{69.3}\\
    \hline
         \multirow{5}{*}{ResNet-50}
         &HSNet~\cite{min2021hsnet}&64.3&70.7&60.3&60.5&64.0&70.3&73.2&67.4&67.1&69.5\\
         &PFENet~\cite{tian2020prior}&61.7&69.5&55.4&56.3&60.8&63.1&70.7&55.8&57.9&61.9\\
         &CyCTR~\cite{zhang2021cycle}&65.7&71.0&59.5&59.7&64.0&69.3&73.5&63.8&63.5&67.5\\
         &SSP~\cite{fan2022self}&60.5&67.8&66.4&51.0&61.4&67.5&72.3&\textbf{75.2}&62.1&69.3\\
         &DCAMA~\cite{shi2022dense}&67.5&72.3&59.6&59.0&64.6&70.5&73.9&63.7&65.8&68.5\\
         &SD-AANet~\cite{zhao2023self}&60.9&70.8&58.4&57.3&61.9&65.5&71.6&62.5&62.3&65.5\\
         &BAM~\cite{lang2022BAM}&69.0&73.6&67.6&61.1&67.8&70.6&75.1&70.8&67.2&70.9\\
         \cline{2-12}
         &\textbf{HDMNet (Ours)}&\textbf{71.0}&\textbf{75.4}&\textbf{68.9}&\textbf{62.1}&\textbf{69.4}&\textbf{71.3}&\textbf{76.2}&71.3&\textbf{68.5}&\textbf{71.8}\\
    \specialrule{\wideline}{0pt}{0pt}
    \hline
    \end{tabular}
    \vspace{-0mm}
    \caption{Performance on \pascal~\cite{shaban2017oneshot} using the mIoU (\%) evaluation metric. Results in bold denote the best performance.}
    \label{tab:pascal_results}
    \vspace{-2mm}
\end{table*}

\subsection{Implementation Details} 
 
HDMNet is built upon the Pytorch~\cite{paszke2019pytorch} framework. All models are trained on $4$ NVIDIA GeForce RTX 3090 GPUs and tested on a single GPU. 
The training augmentations of \pascal~dataset and \coco~dataset follow that of~\cite{tian2020prior} for fair comparisons, including random crop, scale, rotate, blur and flip. 

HDMNet is trained in an episode fashion for $200$ and $50$ epochs on \coco~and \pascal, and the batch sizes are set to $6$ and $4$ respectively. During training, AdmW optimizer is adopted the same as~\cite{zhang2021cycle, min2021hsnet}, and the learning rate is set to $0.0001$. In addition, the weight decay is $0.01$, and the ``poly'' strategy is used to adjust the learning rate.

We use ResNet-50~\cite{He2016resnet} and VGG-16~\cite{simonyan2014very} as the encoder to extract features with freezing parameters to verify the effectiveness of the proposed method on different backbones. PSPNet~\cite{zhao2017pspnet} serves as the base learner in all experiments. We apply the PPM module~\cite{zhao2017pspnet}, which was widely used in previous semantic segmentation methods, to provide multi-resolution context for feature enrichment after the $4th$ block of ResNet-50 or VGG-16 to generate the prior mask~\cite{tian2020prior}. Similar to that of~\cite{tian2020prior}, we concatenate the prior mask and match features with a $1\times1$ kernel size convolution layer leveraging the high-level semantic information to boost performance. During testing, predictions are resized back to the original sizes of the input images, keeping the ground-truth labels intact~\cite{tian2020prior}.

\begin{table}[ht]
     \small
     \centering
     \scalebox{0.9}{
     \begin{tabular}{clccc}
     \specialrule{\wideline}{0pt}{0pt}
     \hline
          \multirow{2}{*}{Backbone}&\multicolumn{1}{l}{\multirow{2}{*}{Methods}}&\multicolumn{2}{c}{FB-IoU (\%)}&\multirow{2}{*}{ \shortstack{\#learnable\\params}}\\
          &&1-shot&5-shot& \\
          \hline
          \multirow{4}{*}{ResNet-50}
          &ASGNet~\cite{li2021adaptive}&60.4&67.0&10.4M\\
          &HSNet~\cite{min2021hsnet}&68.2&70.7&\textbf{2.6M}\\
          &BAM~\cite{lang2022BAM}&71.1&73.3&4.9M\\
          &\textbf{HDMNet (Ours)}&\textbf{72.2}&\textbf{77.7}&4.2M\\

     \specialrule{\wideline}{0pt}{0pt}
     \hline
     \end{tabular}
     }
     \caption{Comparison of results on \coco in terms of FB-IoU and the number of learnable parameters.}
     \label{tab:FBIoU_Params}
     \vspace{-3mm}
 \end{table}

 \begin{table}[ht]
     \small
     \centering
     \scalebox{1.}{
     \begin{tabular}{cccccc}
     \specialrule{\wideline}{0pt}{0pt}
     \hline
          Ens.&HDM&Corr.&Distill&mIoU (\%)&$\Delta$ \\
          \hline
          &&&&44.7&0.0\\
          $\checkmark$&&&&45.8&+1.1\\
          $\checkmark$&$\checkmark$&&&47.9&+3.2\\
          $\checkmark$&$\checkmark$&$\checkmark$&&48.3&+3.6\\
          $\checkmark$&$\checkmark$&$\checkmark$&$\checkmark$&\textbf{50.0}&+\textbf{5.3}\\
     \specialrule{\wideline}{0pt}{0pt}
     \hline
     \end{tabular}
     }
     \caption{Ablation studies for different components and architecture design in HDMNet.}
     \label{tab:ablation_implemetation}
     \vspace{-0.4in}
 \end{table}

\subsection{Comparison with State-of-the-Art Methods}

In Tables \ref{tab:coco_results} and \ref{tab:pascal_results}, we report comparison of our proposed HDMNet with other state-of-the-art few-shot semantic segmentation approaches in recent years on \coco~\cite{nguyen2019featureweight}~and \pascal~\cite{shaban2017oneshot}~datasets. The mIoU (\%) is used as the evaluation metric. To verify the generality, we build our baseline with VGG-16~\cite{simonyan2014very} and Resnet-50~\cite{He2016resnet}. Both of them gain significant improvement from our method, and our method achieves new state-of-the-art performance on both \coco~and \pascal datasets. Especially on \coco~dataset, our model outperforms the prior arts by a significant margin, achieving $4.8$\% (1-shot) and $7.5$\% (5-shot) of mIoU improvements over the SOTA with ResNet-50 backbone. 

\coco~dataset contains $80$ categories compared to $20$ classes in \pascal dataset and has a particularly larger image capacity. The superiority on \coco~dataset proves that our method is with higher generality and can better adapt to novel categories in more complex scenes. Table \ref{tab:pascal_results} shows that our HDMNet also achieves $69.4$\% (1-shot) and $71.8$\% (5-shot) of mIoU on Pascal with ResNet-50 backbone, surpassing previous state-of-the-art. Table~\ref{tab:FBIoU_Params} gives comparison in terms of FB-IoU and model parameter number on~\coco~for 1-shot and 5-shot segmentation. HDMNet achieves the best performance without compromising efficiency.

\subsection{Ablation Study}

We report the ablation study results in this section to investigate the effectiveness of each component and our design choice. All ablation experiments are conducted under~\coco~ 1-shot setting with ResNet-50 backbone if not otherwise specified.

\begin{table}
    \parbox{0.5\linewidth}{
    \centering
         \begin{tabular}{lc}
              \specialrule{\wideline}{0pt}{0pt}
              \hline
              Matching&mIoU(\%)\\
              \hline
              CA&48.0\\
              Cos&49.2\\
              Cos+SM&49.5\\
              Cos+Inv-SM&\bf{50.0}\\
         \specialrule{\wideline}{0pt}{0pt}
         \hline
         \end{tabular}
    \caption{Ablation study on correlation mechanism. CA: Cross-Attention. Cos: Cosine similarity. Inv-SM: Inversed Softmax.}
    \label{tab:correlation_module}
    }
    \hfill
    \parbox{0.43\linewidth}{
    \centering
         \begin{tabular}{lc}
         \specialrule{\wideline}{0pt}{0pt}
         \hline
              Loss&mIoU(\%)\\
              \hline
              w/o&47.9\\
              CE&48.4\\
              CE+KD&49.1\\
              KL+KD&\bf{50.0}\\
         \specialrule{\wideline}{0pt}{0pt}
         \hline
         \end{tabular}
    \caption{Ablation study on the different loss functions. KD indicates using
    the adjacent layers' soft predictions.}
    \label{tab:loss_function}
    }
    \vspace{-.1in}
\end{table}

\vspace{-0.1in}
\paragraph{Component-wise ablation.} Table~\ref{tab:ablation_implemetation} shows ablation results regarding the effectiveness of different components and architecture design, where the mIoU results are averaged over four splits. 

The first line is the baseline result. We build our baseline following~\cite{tian2020prior,zhang2021cycle} and utilize ResNet-50~\cite{He2016resnet} as the encoder to extract image features and generate the prior mask~\cite{tian2020prior}. Extracted features from Block-2 and Block-3 of the backbone are fused and fed to the next step with the prior mask. The baseline stacks the self-attention and cross-attention modules three times and applies the cross-attention module for feature matching~\cite{zhang2021cycle}. 

For a fair comparison, we use the ensemble module (Ens.) following BAM to filter the categories appearing in the training process. The test result is incrementally improved. Then we convert the framework to our proposed structure (HDM) described in Sec. \ref{sec:MSSMStructure}, resulting in even better results, with further mIoU increase of $2.1\%$ compared to the one with Ens. only.
It indicates that decoupling feature down-sampling and matching reduce noise interference and boost performance. Replacing the attention function with the proposed correlation mechanism (Corr.) as mentioned in~\secref{sec:correlation_mechanism}~within the matching module continuously improves the performance by $0.4\%$ mIoU.

Finally, we use the correlation map distillation (Distill) described in~\secref{sec:correlation_distillation} to supervise the matching process and leverage the correlation information from different stages, bringing another $1.7\%$ mIoU improvement, which validates the effectiveness of our proposed distillation strategy. 

\vspace{-0.1in}
\paragraph{Correlation mechanism ablation.} 
Table \ref{tab:correlation_module} compares the alternative choices of the correlation mechanism. We first build $\text{HDMNet}$ with the original cross-attention layer (CA) ~\cite{vaswani2017attention} as the matching module and apply the distillation function to the attention map, \ie, dot product between the query and key sequences. Then we replace the matching module with the cosine similarity (Cos) and inverse softmax (Inv-SM) as described in~\secref{sec:correlation_mechanism}. The results prove that the proposed correlation mechanism is conducive to the final performance.

\begin{table}
     \small
     \centering
     \scalebox{.93}{
     \begin{tabular}{c|cccc}
     \specialrule{\wideline}{0pt}{0pt}
     \hline
     Decoder&mIoU(\%)&{params(\textit{M})}&time(\textit{ms})&FLOPs(\textit{G})\\
        \hline
        CyCTR~\cite{zhang2021cycle}&40.3&5.6&54.3&96.7\\
        HSNet~\cite{min2021hsnet}&39.2&2.6&25.5&20.6\\
        BAM~\cite{lang2022BAM}&45.2&4.1&\bf{7.4}&26.0\\
        \hline
        Ours-$S_1$&47.1&\bf{1.3}&15.1&\bf{8.8}\\
        Ours-$S_2$&48.8&2.1&21.1&10.2\\
        Ours-$S_3$&\bf{50.0}&2.8&27.4&10.6\\
        Ours-$S_4$&48.6&3.6&38.0&10.6\\
     \specialrule{\wideline}{0pt}{0pt}
     \hline
     \end{tabular}
     }
     \caption{Comparison of decoders from different methods in terms of accuracy, efficiency, and model size. $S_i$ indicates constructing our decoder with $i$ matching stages. }
     \label{tab:ablation_levels}
 \end{table}
 
 \begin{figure}
     \centering
     \scalebox{1}[1]{\includegraphics[width=0.46\textwidth]{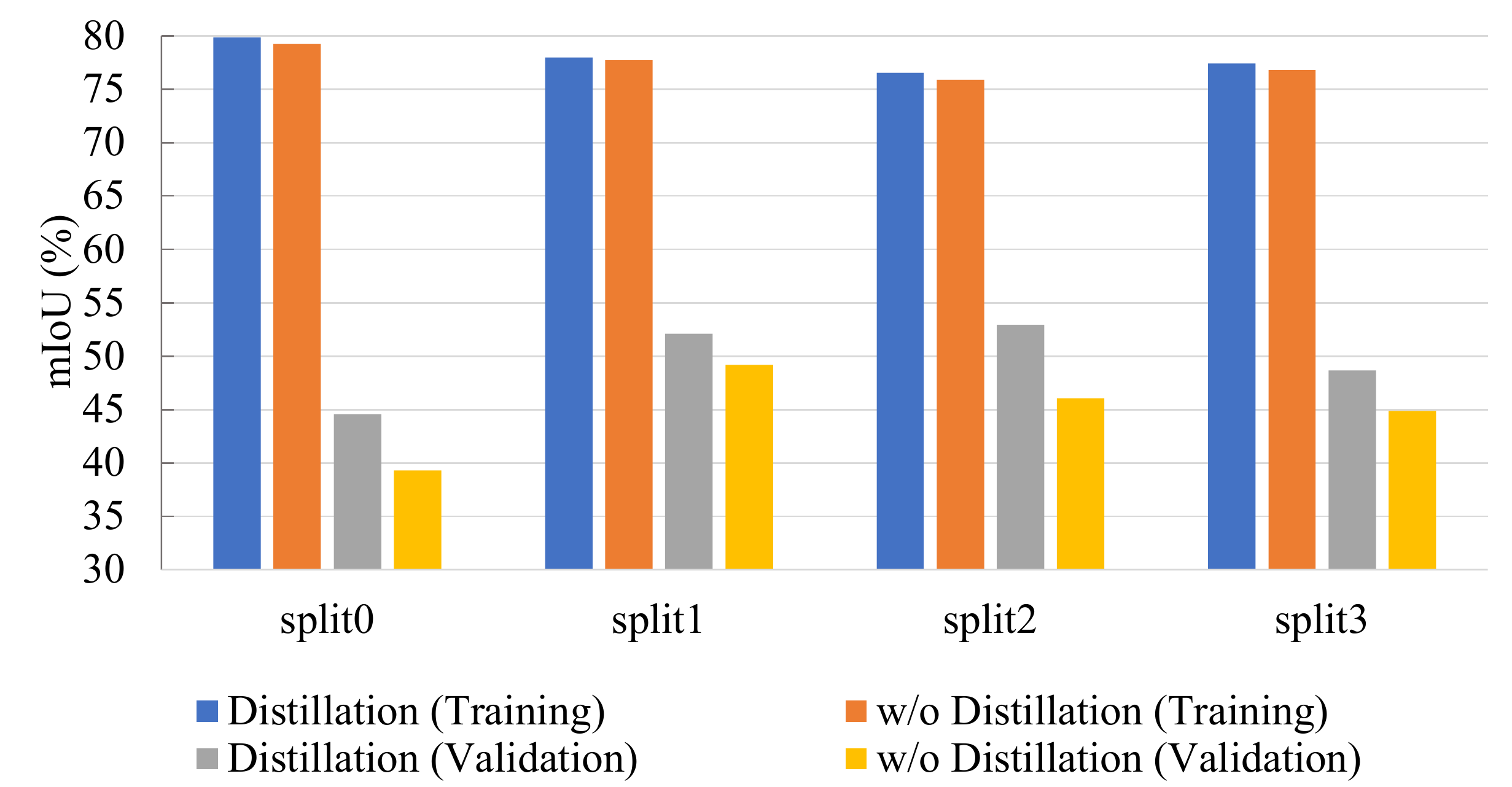}}
     \caption{Ablation study on correlation map distillation for training and validation on \coco~dataset in terms of mIoU. }
     \vspace{-.2in}
     \label{fig:ablation_ditillation_chart}
 \end{figure}  

\vspace{-0.2in}
\paragraph{Distillation ablation.} To verify the effectiveness of correlation map distillation, we set up two controlled experiments under \coco~1-shot setting.~\figref{fig:correlation_visualization} shows the activation maps of correlation. 

\figref{fig:ablation_ditillation_chart}~plots the training and validation results with or without correlation map distillation on every \coco~split. The correlation map distillation has limited influence on the training process. But it largely improves the validation results.

Finally, we compare different loss functions in Table~\ref{tab:loss_function}. We directly adopt the cross entropy (CE) loss between the correlation maps and the one-hot labels in ground-truth annotations and then use knowledge distillation (KD) by selecting adjacent layers as the teacher and student. KL denotes the Kullback-Leibler divergence loss.

\begin{figure}
     \centering
     \includegraphics[width=0.48\textwidth]{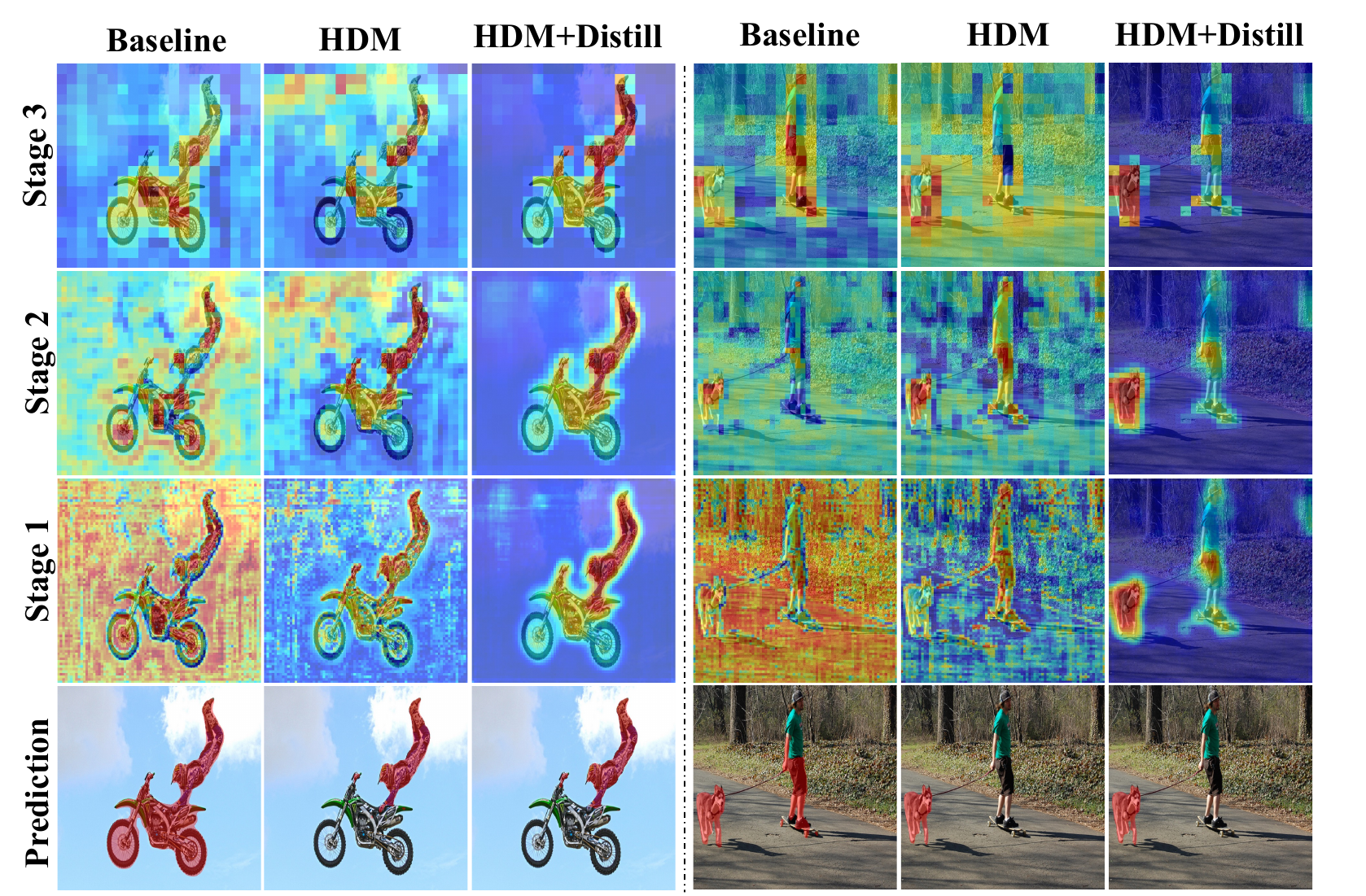}
     \caption{Qualitative correlation maps in 1-3 matching stages. The target classes of the left and right panels are ``people'' and ``dog'' respectively.}
     \label{fig:ablation_pyramid_level}
     \vspace{-.2in}
 \end{figure}

\vspace{-0.1in}
\paragraph{Effect of the matching pyramid.} Table \ref{tab:ablation_levels} compares the decoders of previous methods and our proposed matching pyramid with different stage numbers in terms of accuracy, efficiency, and model size.~\figref{fig:ablation_pyramid_level} visualizes qualitative results of correlation maps in 1-3 matching stages under distinct designs. We build our baseline by directly interleaving the self- and cross-attention layers and then convert it to the hierarchically decoupled matching (HDM) structure. HDMNet  better fights against the interference of other classes in the same image but suffers from mining the correlation information from the shallow stages,  which contain more detailed features but fewer semantic cues. We finally adopt correlation map distillation to facilitate the earlier layers to be more aware of the contextual information.

\vspace{-0.1in}
\section{Concluding Remarks}
\label{sec:con}
In this paper, we have proposed hierarchically decoupled matching network (HDMNet) to tackle the challenging few-shot segmentation problem. HDMNet decouples the downsampling and matching process to prevent information interference. Further, we designed a novel matching module constructed on the correlation mechanism and distillation and performed extensive experiments to demonstrate that this design alleviates the training-class overfitting problem and improves generality. 
One limitation is that our model sometimes may fail to clearly distinguish different categories that have rather similar appearances and close semantic relations, such as truck and car, chair and sofa. It may be a promising direction for future research.

\vspace{-.1in}
\paragraph{Acknowledgements.} This work is partially supported by ITF Partnership Research Programme (No.PRP/65/20FX) and Shenzhen Science and Technology Program KQTD20210811090149095.


{\small
\bibliographystyle{ieee_fullname}
\bibliography{egbib}
}

\clearpage
\appendix
\section*{Appendix}

\section{Datasets}
Compared with the previous work, we mainly use \coco~\cite{nguyen2019featureweight} for verification in our experiments, which contains a more extensive data size. In this section, we give more research and analysis to prove that when the data is too clean and straightforward, few-shot segmentation will degenerate into foreground segmentation without caring about specific semantic information, and we conclude that \coco~dataset will be a better choice to verify the model's generality.

\subsection{Statistical Analysis} 
\begin{figure}[!ht]
   \centering
   \includegraphics[width=0.45\textwidth]{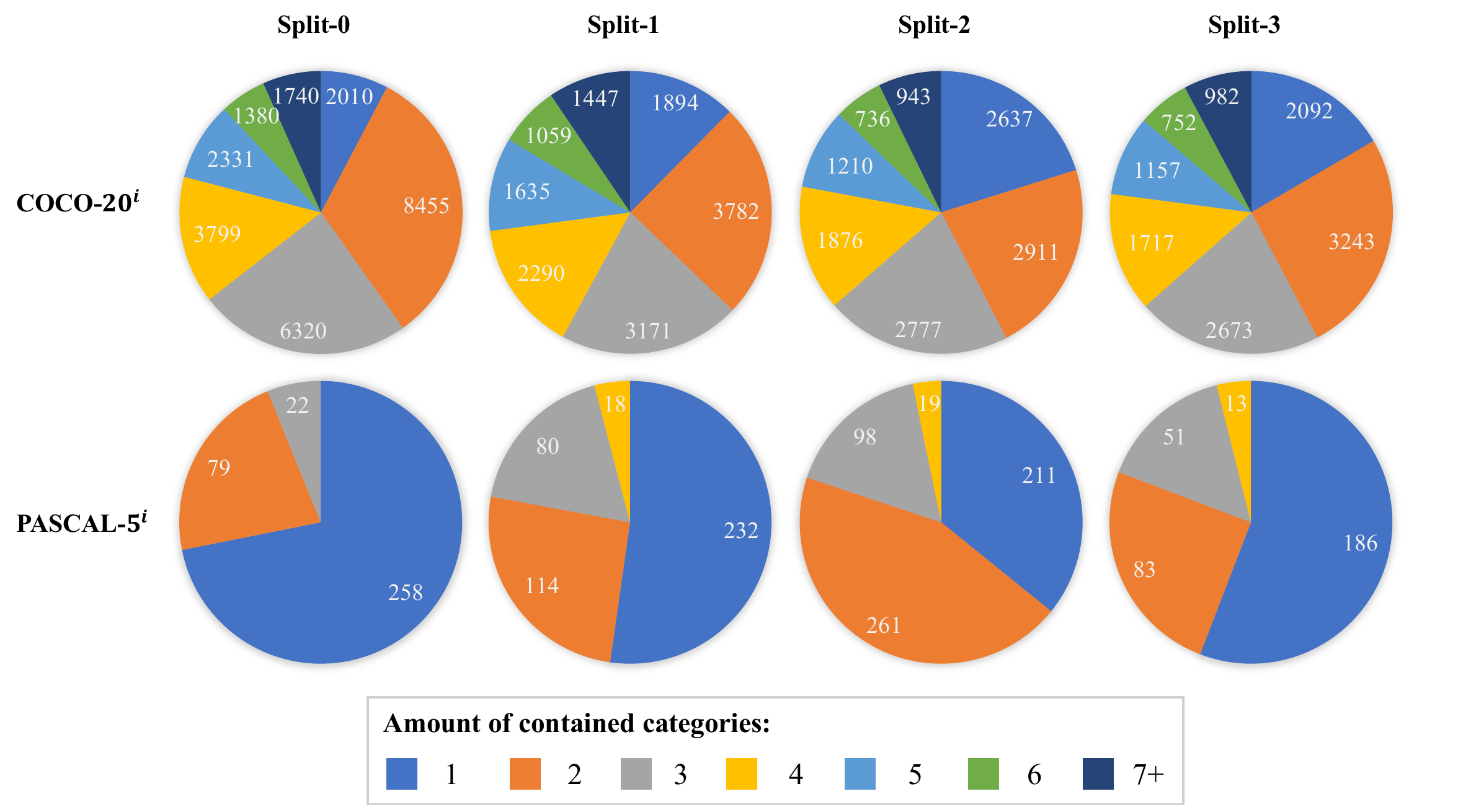}
   \caption{Statistics of pictures containing different category amounts for each fold of \coco~\cite{nguyen2019featureweight} and \pascal~\cite{shaban2017oneshot} datasets.}
   \label{fig:statistic_datasets}
\end{figure}

The performance gap between the \coco~\cite{nguyen2019featureweight}and \pascal~\cite{shaban2017oneshot} datasets was attributed to the class amount and data quantity. However, we found that the image's complexity is also a key factor. We first count the number of pictures with different contained category amounts, and the results are shown in ~\figref{fig:statistic_datasets}. It can be found that most images of \pascal have only single foreground. The model only needs to distinguish the significant foreground without semantic support, and few-shot segmentation will degenerate into foreground segmentation tasks.

\subsection{Ablation Experiment of the Support Mask}
\begin{figure*}[!ht]
   \centering
   \includegraphics[width=0.7\textwidth]{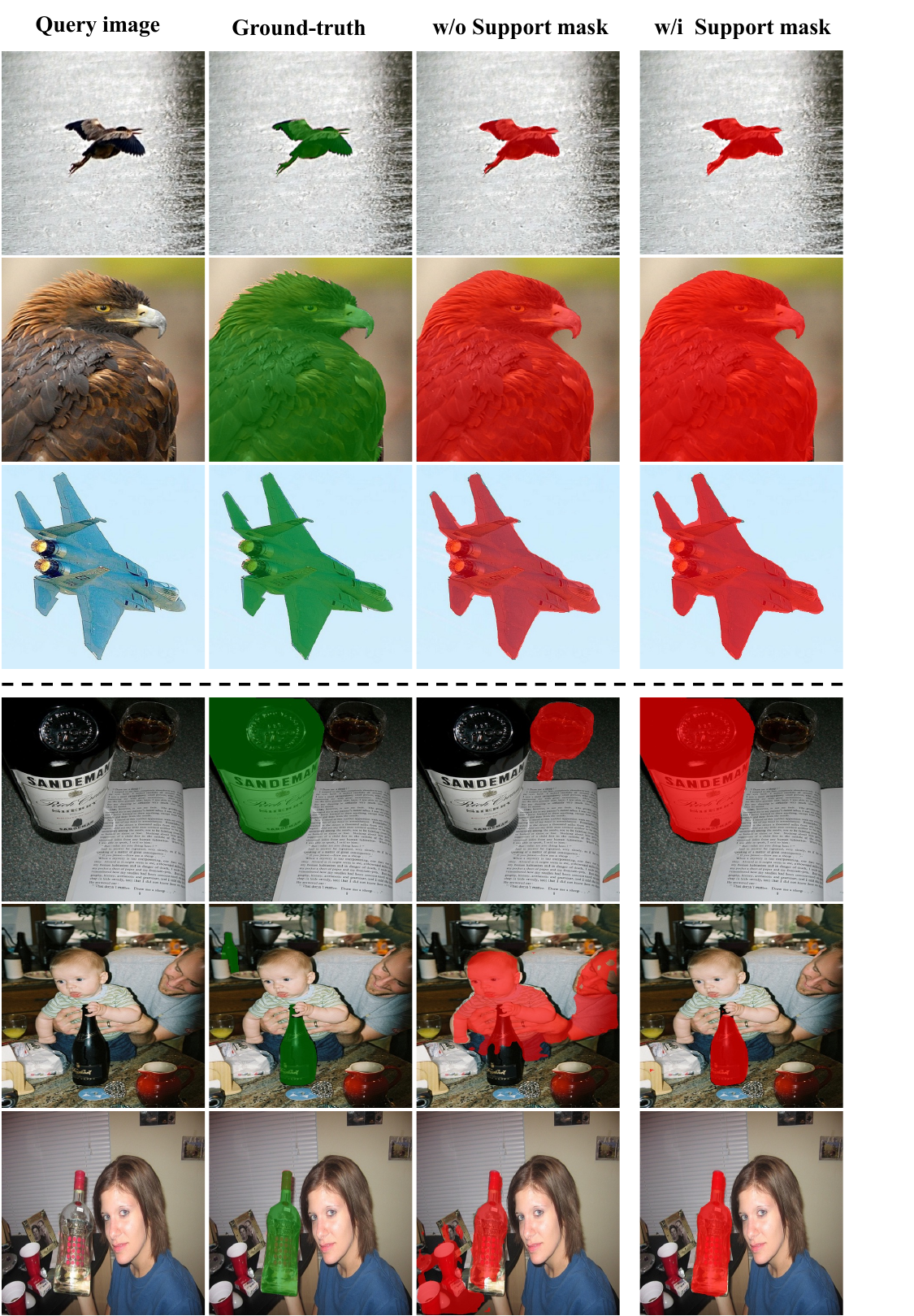}
   \caption{Qualitative results of the ablation studies for the support mask's effects under \pascal~\cite{shaban2017oneshot} $1$-hot setting. The upper panel shows the query images with only single foreground, while the query images from the below panel contain complex scenes.}
   \label{fig:qualitative_ablation_Ms}. 
\end{figure*}

We also ingeniously designed a simple ablation experiment to verify whether the model can extract the supervision information from the support annotations. First, we conduct the experiment following the original few-shot semantic segmentation settings~\cite{tian2020prior}, and then we give the model inputs without support supervision by removing the support mask. Tab.~\ref{tab:abblation_support_mask} and ~\figref{fig:qualitative_ablation_Ms}~showthe quantized and qualitative results, respectively. It is noteworthy that the model also can achieve stunning results on \pascal~dataset, even without the support mask. In \figref{fig:qualitative_ablation_Ms}, we visualize the performances under the simple and complex scenes. When the images are too straightforward, containing only single foreground with a clean background, few-shot segmentation will degenerate into foreground segmentation and achieve incredible performance even without caring about semantic information. 

\begin{table}[ht]
   \small
   \centering
   \scalebox{0.8}{
   \begin{tabular}{clcccccc}
   \specialrule{\wideline}{0pt}{0pt}
   \hline
        \multirow{2}{*}{Dataset}&\multirow{2}{*}{Method}&\multicolumn{5}{c}{mIoU(\%)}\\
        &&Fold-0&Fold-1&Fold-2&Fold-3&Mean\\
        \hline
        \multirow{2}{*}{$\text{PASCAL-}5^i$}
        &w/ $\mM^s$&71.2&75.4&67.6&63.6&69.5\\
        &w/o $\mM^s$&62.8&69.4&59.3&53.6&61.3\\
        \hline
        \multirow{2}{*}{$\text{COCO-}20^i$}
        &w/ $\mM^s$&43.8&55.3&51.6&49.4&50.0\\
        &w/o $\mM^s$&31.0&37.7&31.2&33.8&33.4\\
   \specialrule{\wideline}{0pt}{0pt}
   \hline
   \end{tabular}
   }
   \caption{Ablation studies of the support mask's effects for \pascal~\cite{shaban2017oneshot} and \coco~\cite{nguyen2019featureweight} datasets under $1$-hot setting.}
   \vspace{-.1in}
   \label{tab:abblation_support_mask}
\end{table}

\section{Number of Test Episodes}

Few-shot semantic segmentation adopts episode paradigm testing, where each test episode randomly selects the query and support pairs containing the same class objects. Typically, $1k$ episodes are set in the \pascal~evaluation since there are at most $584$ images in each pascal fold. In contrast, we set $10k$ random episodes for each fold evaluation on \coco~dataset and calculate the average mIoU. In this work, we give more experiments results proving that only $1k$ episodes are not sufficient to provide reliable results on \coco~for comparison. 

\begin{figure}[ht]
   \centering
   \includegraphics[width=.47\textwidth]{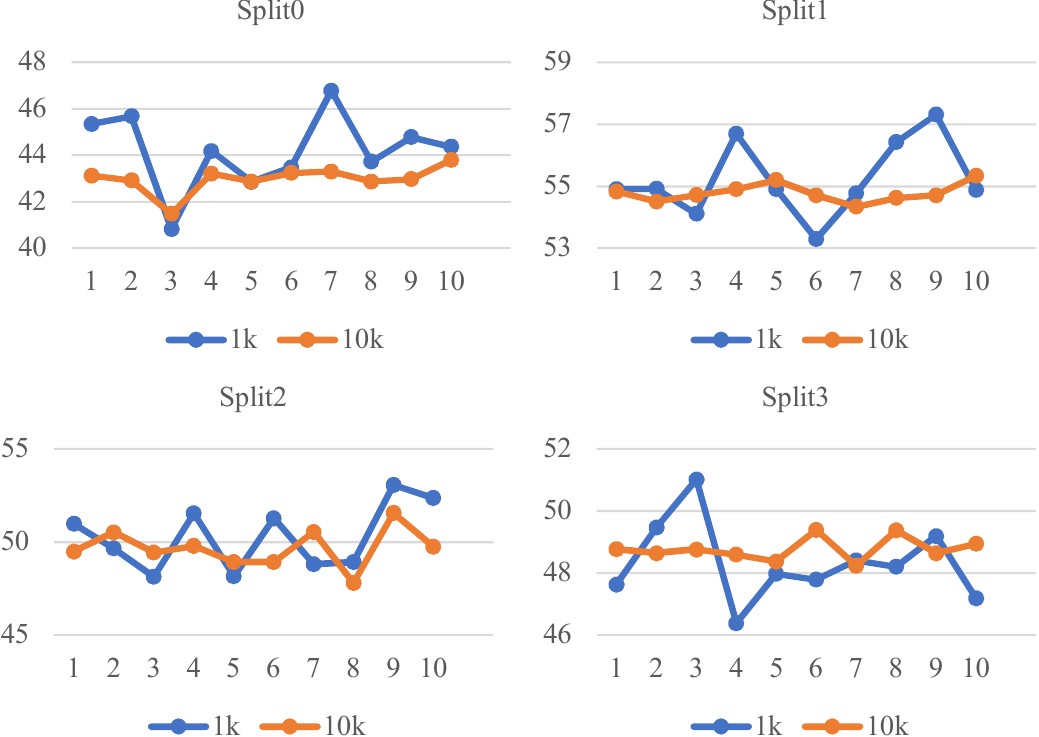}
   \caption{Comparison between $1k$ and $10k$ test samples set on \coco~dataset for each fold.}
   \label{fig:iteration_comparison}
\end{figure}

We iterate $10$ times evaluations and then plot the line chart of each fold's result for both $1k$ and $10k$ episodes set as shown in \figref{fig:iteration_comparison}. Insufficiency sampling episodes will lead to a precarious and significant fluctuation of test results. For example, the difference between the results is up to $6.0\%$ mIoU in \coco~Fold-0.

\section{Implemetation Details}

\subsection{Decoder Structures}
\begin{figure}[!ht]
   \centering
   \includegraphics[width=0.45\textwidth]{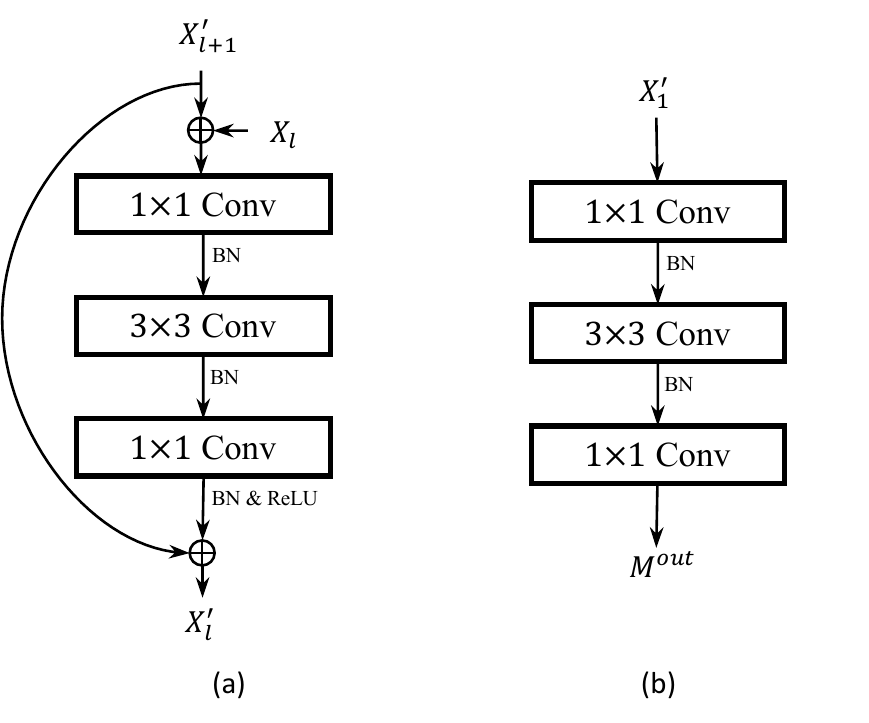}
   \caption{Structures of (a) decoder block and (b) classification head. Specifically, we adopt batch normalization (BN) on single GPU machine and convert it to sync batch normalization (SyncBN) for multi-GPU training. }
   \label{fig:module_arch}
   \vspace{-.1in}
\end{figure}

The decoder fuses the matching results $\{\mX_l\in\R^{c_l\times h_l^q\times w_l^q}\}_{l=1}^L$ from coarse resolution to fine grain. The structure of decoder block and classification head is illustrated in~\figref{fig:module_arch}. Decoder block inputs feature-match results from the same stage and sequential output from the last stage block following hierarchical paradigm. We use residual connection~\cite{He2016resnet} to alleviate exploding/vanishing gradient problem. Classification head inputs the last stage output with maximum resolution and predicts the dense mask as the final output.

\subsection{Experimental Environment}
Software and hardware environment:
\begin{itemize}
   \item CUDA version: 11.7
   \item PyTorch version: 1.12.1
   \item GPU: NVIDIA GeForce RTX 3090
   \item CPU: Intel(R) Xeon(R) Gold 6326 CPU @ 2.90GHz 
\end{itemize}

\subsection{Additional Qualitative  Results}
\begin{figure*}[!ht]
   \centering
   \includegraphics[width=0.9\textwidth]{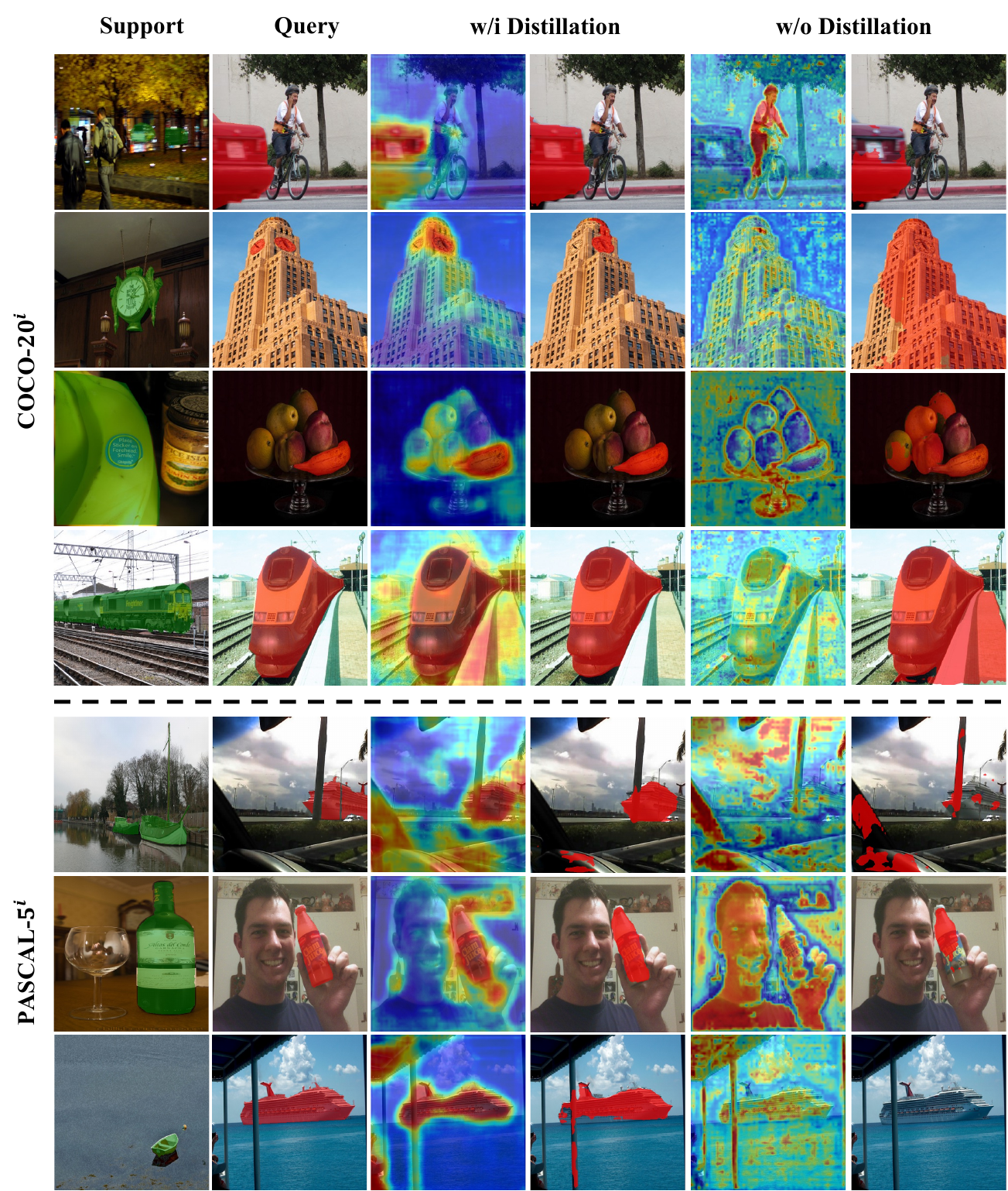}
   \caption{More visualization results of the correlation maps on both \pascal~\cite{shaban2017oneshot} and \coco~\cite{nguyen2019featureweight} under 1-shot setting. The first and second columns show examples of the support images with ground truth in green and the query images with labeled masks in red, respectively. Then we show the correlation maps and prediction results within or without distillation, respectively. We select the correlation map from the first stage for a brief introduction and visualize them by heatmaps.}
   \label{fig:distillation_heatmap}
\end{figure*}

\begin{figure*}[!ht]  
   \centering
   \includegraphics[width=1.18\textwidth, angle=-90]{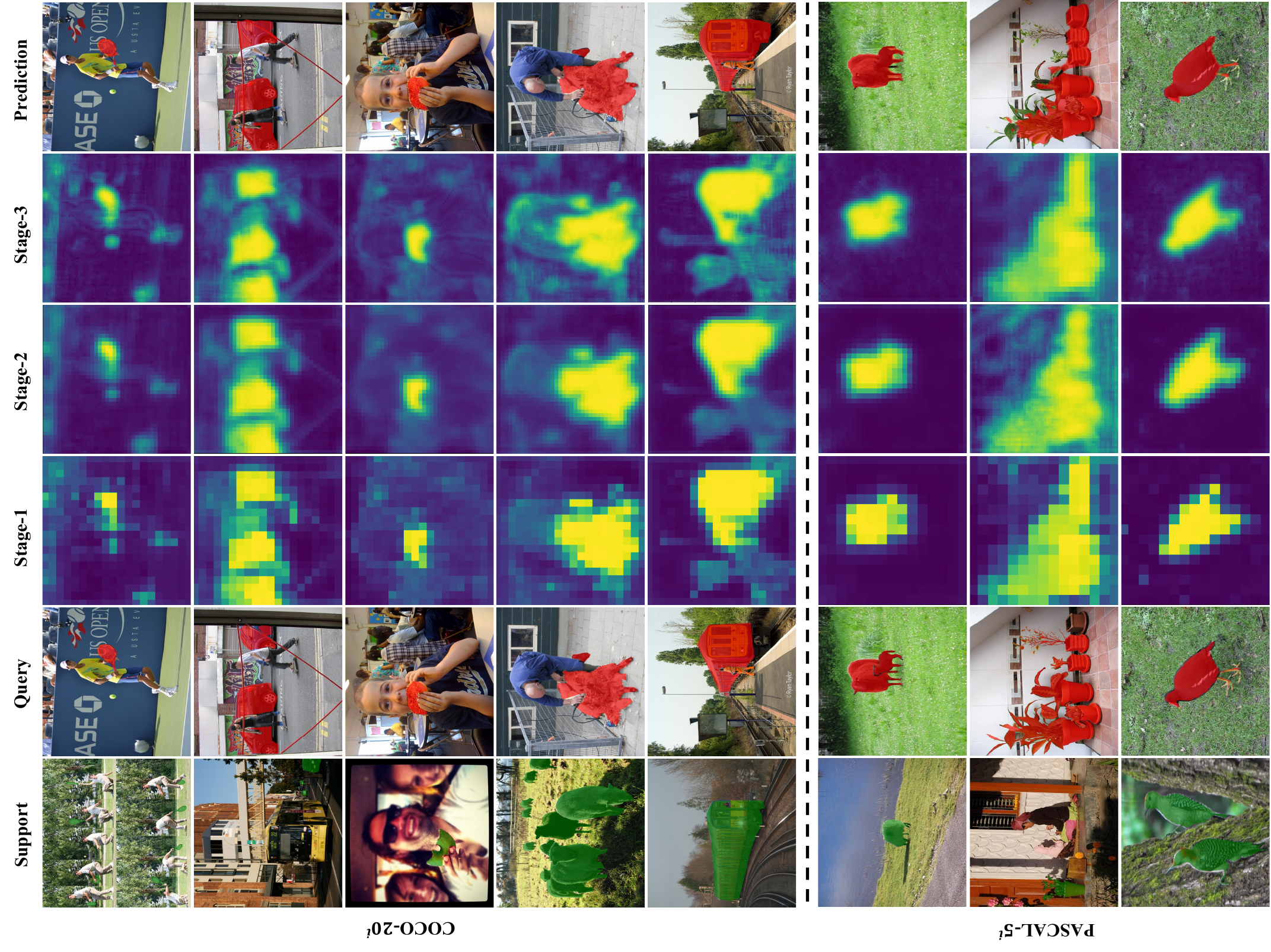}
   \caption{More qualitative results of the correlation pyramid on both \pascal~\cite{shaban2017oneshot} and \coco~\cite{nguyen2019featureweight} under 1-shot setting. The first and second columns show examples of the support images with ground truth in green and the query images with labeled masks in red, respectively. The following three
   columns visualize the correlation pyramid from the first to the third
   stage of the matching pyramid and the last column is the model's outputs.}
   \label{fig:pyramid_stage}
\end{figure*}

More qualitative results are provided to validate and analyze our proposed network effectiveness.~\figref{fig:distillation_heatmap} visualizes correlation maps and compared the dense predicted mask with or without correlation distillation on both \pascal~\cite{shaban2017oneshot} and \coco~\cite{nguyen2019featureweight} under 1-shot setting.~\figref{fig:pyramid_stage} shows the correlation maps from 1-3 pyramid stages. The correlation maps from the coarse resolution give rough locations of the classes related to the support annotations, and the fine layers provide more detailed features facilitating segmentation.

\section{Ablation Experiments} 

\subsection{Distillation Temperature}
During the distillation process, we adopt the temperature T, a hyperparameter, to control the distribution. Given flattened correlation maps, we first apply a softmax layer with T to perform the spatial normalization among all positions:
\begin{equation}
   \hat{\mC}'_l(i)=\frac{\text{exp}({\mC}'_l(i)/T)}{\sum_{j=1}^{h_l^qw_l^q}\text{exp}({\mC}'_l(j)/T)},
   \label{eq:softmax_normalization2}
\end{equation}
where $l$ indicates the stage, $T$ denotes the temperature of distillation~\cite{hinton2015distilling}. In this section, we study the influence of the temperature set through the ablation experiments, and all results are shown in Tab.~\ref{tab:abblation_hyperparameter_T}. When $T$ equals 1, we get the best performance, and we keep this through all experiments as a default if not otherwise specified. 
\begin{table}[ht]
    \small
    \centering
    \scalebox{.99}{
    \begin{tabular}{c|ccccc}
    \specialrule{\wideline}{0pt}{0pt}
    \hline
         Temperature&\multicolumn{5}{c}{mIoU(\%)}\\
         ( $T$ )&Fold-0&Fold-1&Fold-2&Fold-3&Mean\\
         \hline
         0.5&42.1&54.9&48.6&48.6&48.6\\
         1&\bf{43.8}&\bf{55.3}&\bf{51.6}&49.4&\bf{50.0}\\
         2&43.1&55.1&50.0&\bf{49.5}&49.4\\
         5&42.4&54.9&48.2&47.5&48.3\\
    \specialrule{\wideline}{0pt}{0pt}
    \hline
    \end{tabular}
    }
    \caption{Ablation studies of the distillation temperature.}
    \label{tab:abblation_hyperparameter_T}
 \end{table}

\end{document}

%% file: res/shortcuts.tex
\usepackage{times}
\usepackage{epsfig}
\usepackage{graphicx}
\usepackage{wrapfig}
\usepackage{amsmath,amssymb,amsthm}
\usepackage{algorithm,algorithmicx,algpseudocode}
\usepackage{bm,xspace}
\usepackage{comment}
\usepackage{verbatim}
\usepackage{multirow}
\usepackage{balance}
\usepackage{url}
\usepackage{booktabs}
\usepackage{etoolbox,siunitx}
\usepackage{calc}
\usepackage{pifont,hologo}
\usepackage{color}

\newcommand{\figref}[1]{Fig.~\ref{#1}}

\newcommand{\secref}[1]{Sec.~\ref{#1}}

\usepackage[super]{nth}
\usepackage{nicefrac}
\robustify\bfseries

\def\mM{\mathcal{M}}

\def\pascal{$\text{PASCAL-}5^i$}
\def\coco{$\text{COCO-}20^i$}
\def\wideline{0.6pt}

\DeclareMathSymbol{@}{\mathord}{letters}{"3B}

\def\latex/{\LaTeX}
\def\bibtex/{\hologo{BibTeX}}

\makeatletter
\DeclareRobustCommand\onedot{\futurelet\@let@token\@onedot}
\def\@onedot{\ifx\@let@token.\else.\null\fi\xspace}
 
\def\ie{\emph{i.e}\onedot}

%% file: res/math_commands.tex
\newcommand*{\Rom}[1]{\expandafter\@slowromancap\romannumeral #1@}
\newcommand*{\rom}[1]{\expandafter\romannumeral #1}

\def\1{\bm{1}}

\def\mC{{\bm{C}}}
\def\mD{{\bm{D}}}

\def\mF{{\bm{F}}}

\def\mI{{\bm{I}}}

\def\mK{{\bm{K}}}

\def\mM{{\bm{M}}}

\def\mQ{{\bm{Q}}}

\def\mS{{\bm{S}}}

\def\mV{{\bm{V}}}
\def\mW{{\bm{W}}}
\def\mX{{\bm{X}}}

\DeclareMathAlphabet{\mathsfit}{\encodingdefault}{\sfdefault}{m}{sl}
\SetMathAlphabet{\mathsfit}{bold}{\encodingdefault}{\sfdefault}{bx}{n}

\newcommand{\R}{\mathbb{R}}

